# Singular curves and cusp points in the joint space of 3-*RPR* parallel manipulators

Mazen ZEIN , Philippe WENGER and Damien CHABLAT
Institut de Recherche en Communications et Cybernétique de Nantes UMR CNRS 6597
1, rue de la Noe, BP 92101, 44321 Nantes Cedex 03 France
Mazen.Zein@irccyn.ec-nantes.fr

*Abstract*— This paper investigates the singular curves in two-dimensional slices of the joint space of a family of planar parallel manipulators. It focuses on special points, referred to as cusp points, which may appear on these curves. Cusp points play an important role in the kinematic behavior of parallel manipulators since they make possible a nonsingular change of assembly mode. The purpose of this study is twofold. First, it reviews an important previous work, which, to the authors' knowledge, has never been exploited yet. Second, it determines the cusp points in any two-dimensional slice of the joint space. First results show that the number of cusp points may vary from zero to eight. This work finds applications in both design and trajectory planning.

## I. INTRODUCTION

Because at a singularity a parallel manipulator loses stiffness, it is of primary importance to be able to characterize these special configurations. This is, however, a very challenging task for a general parallel manipulator [1]. Planar parallel manipulators have received a lot of attention [1-8] because of their relative simplicity with respect to their spatial counterparts. Moreover, studying the formers may help understand the latters. Planar manipulators with three extensible leg rods, referred to as 3-*RPR* manipulators, have often been studied. Such manipulators may have up to six assembly modes [7]. The direct kinematics can be written in a polynomial of degree six [3]. Moreover, as is the case in most parallel manipulators, the singularities coincide with the set of configurations where two direct kinematic solutions coincide. It was first pointed out that to move from one assembly mode to another, the manipulator must cross a singularity [7]. But [8] showed, using numerical experiments, that this statement is not true in general. More precisely, this statement is true only under some special geometric conditions such as similar base and mobile platforms [9,10]. In fact, an analogous phenomenon exists in serial manipulators that can change their posture (inverse kinematic solution) without meeting a singularity in general, but not under special geometric simplification [8,11]. The nonsingular change of posture in serial manipulators was shown to be associated with the existence of points in the workspace where three inverse kinematic solutions meet, called cusp points [11]. Likewise, [9] pointed out that for a 3-*RPR* parallel manipulators (as well as for its spatial counterpart, the octahedral manipulator), if a point with triple direct kinematic solutions exists in the joint space, then the nonsingular change of assembly mode is possible. A condition for three direct kinematic solutions to coincide was established. However, no systematic exploitation of this condition was possible because the algebra involved was too complicated and to the authors' knowledge, the work of [9] has never been pursued yet.

In this paper, the abovementioned condition is reviewed and exploited. An algorithm for the systematic detection of cusp points is developed. It is shown that the number of cusp points depends on the slice of the joint space in which these cusp points are determined and the maximum number of cusp points depends on the geometry of the manipulator. This work helps better understand the topology of the joint space of parallel manipulators and finds applications in both design and trajectory planning.

The following section introduces the 3-*RPR* manipulator and its constraint equations. Sections III is devoted to the determination of the singular curves in slices of the joint space. The existence condition of cusp points is derived in section IV and an algorithm to automatically determine these points is briefly described. Section V is devoted to the explanation of the results obtained.

## II. PRELIMINARIES

### A. Manipulators Under Study

The manipulators under study are 3-DOF planar parallel manipulators with three extensible leg rods (Fig. 1).

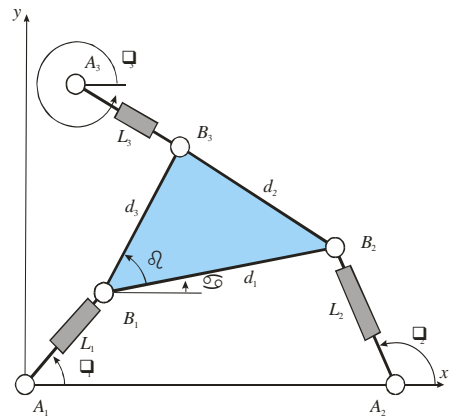

Figure 1. The 3-*RPR* parallel manipulator under study.

This manipulator has been frequently studied [4-8]. Each of the three extensible leg rods is actuated with a prismatic joint. The geometric parameters of the manipulators are the three sides of the moving platform $d_1$, $d_2$, $d_3$ and the position of the base revolute joint centers defined by $A_1$, $A_2$ and $A_3$. The reference frame is centered at $A_1$ and the x-axis passes through $A_2$. Thus, $A_1 = (0, 0)$, $A_2 = (A_{2x}, 0)$ and $A_3 = (A_{3x}, A_{3y})$.

*B. Constraint Equations*

Let $\mathbf{L} \equiv (L_1, L_2, L_3)$ define the lengths of the three leg rods and let $\boldsymbol{\theta} \equiv (\theta_1, \theta_2, \theta_3)$ define the three angles between the leg rods and the x-axis. The six parameters $(\mathbf{L}, \boldsymbol{\theta})$ can be regarded as a configuration of the manipulator but only three of them are independent, so that the configuration space is a 3-dimensional manifold embedded in a 6-dimensional space. The dependency between $(\mathbf{L}, \boldsymbol{\theta})$ can be identified by writing the fixed distances between the three vertices of the mobile platform $B_1$, $B_2$, $B_3$, which yield the following constraint equations

$$\begin{cases} \Gamma_1(\mathbf{L},\boldsymbol{\theta}) = [\mathbf{b}_2(\mathbf{L},\boldsymbol{\theta}) - \mathbf{b}_1(\mathbf{L},\boldsymbol{\theta})]^T [\mathbf{b}_2(\mathbf{L},\boldsymbol{\theta}) - \mathbf{b}_1(\mathbf{L},\boldsymbol{\theta})] - d_1^2 = 0 \\ \Gamma_2(\mathbf{L},\boldsymbol{\theta}) = [\mathbf{b}_3(\mathbf{L},\boldsymbol{\theta}) - \mathbf{b}_2(\mathbf{L},\boldsymbol{\theta})]^T [\mathbf{b}_3(\mathbf{L},\boldsymbol{\theta}) - \mathbf{b}_2(\mathbf{L},\boldsymbol{\theta})] - d_2^2 = 0 \\ \Gamma_3(\mathbf{L},\boldsymbol{\theta}) = [\mathbf{b}_1(\mathbf{L},\boldsymbol{\theta}) - \mathbf{b}_3(\mathbf{L},\boldsymbol{\theta})]^T [\mathbf{b}_1(\mathbf{L},\boldsymbol{\theta}) - \mathbf{b}_3(\mathbf{L},\boldsymbol{\theta})] - d_3^2 = 0 \end{cases}$$
(1)

where $\mathbf{b}_i$ is the vector defining the coordinates of $B_i$ in the reference frame as function of $\mathbf{L}$ and $\boldsymbol{\theta}$. For more simplicity, $(\mathbf{L}, \boldsymbol{\theta})$ will be omitted in the following equations.

Expanding each $\Gamma_i$ as a series about the configuration $(\mathbf{L}, \boldsymbol{\theta})$ yields

$$\Delta\Gamma_i = \left( \sum_{j=1}^{3} \Delta\theta_j \frac{\partial}{\partial \theta_j} + \sum_{j=1}^{3} \Delta L_j \frac{\partial}{\partial L_j} \right) \Gamma_i +$$
$$\frac{1}{2!} \left( \sum_{j=1}^{3} \Delta\theta_j \frac{\partial}{\partial \theta_j} + \sum_{j=1}^{3} \Delta L_j \frac{\partial}{\partial L_j} \right)^2 \Gamma_i + ... +$$
(2)
$$\frac{1}{n!} \left( \sum_{j=1}^{3} \Delta\theta_j \frac{\partial}{\partial \theta_j} + \sum_{j=1}^{3} \Delta L_j \frac{\partial}{\partial L_j} \right)^n \Gamma_i + .. = 0$$

If one keeps only the first-order and second-order terms, Eq. (2) can be written in matrix form as follows

$$\Delta\Gamma = \frac{\partial \Gamma}{\partial \boldsymbol{\theta}} \Delta\boldsymbol{\theta} + \frac{\partial \Gamma}{\partial \mathbf{L}} \Delta\mathbf{L}$$

$$+ \frac{1}{2} \begin{bmatrix} \Delta\boldsymbol{\theta}^T \frac{\partial^2 \Gamma_1}{\partial \boldsymbol{\theta}^2} \Delta\boldsymbol{\theta} \\ \Delta\boldsymbol{\theta}^T \frac{\partial^2 \Gamma_2}{\partial \boldsymbol{\theta}^2} \Delta\boldsymbol{\theta} \\ \Delta\boldsymbol{\theta}^T \frac{\partial^2 \Gamma_3}{\partial \boldsymbol{\theta}^2} \Delta\boldsymbol{\theta} \end{bmatrix} + \begin{bmatrix} \Delta\boldsymbol{\theta}^T \frac{\partial^2 \Gamma_1}{\partial \boldsymbol{\theta} \partial \mathbf{L}} \Delta\mathbf{L} \\ \Delta\boldsymbol{\theta}^T \frac{\partial^2 \Gamma_2}{\partial \boldsymbol{\theta} \partial \mathbf{L}} \Delta\mathbf{L} \\ \Delta\boldsymbol{\theta}^T \frac{\partial^2 \Gamma_3}{\partial \boldsymbol{\theta} \partial \mathbf{L}} \Delta\mathbf{L} \end{bmatrix} + \frac{1}{2} \begin{bmatrix} \Delta\mathbf{L}^T \frac{\partial^2 \Gamma_1}{\partial \mathbf{L}^2} \Delta\mathbf{L} \\ \Delta\mathbf{L}^T \frac{\partial^2 \Gamma_2}{\partial \mathbf{L}^2} \Delta\mathbf{L} \\ \Delta\mathbf{L}^T \frac{\partial^2 \Gamma_3}{\partial \mathbf{L}^2} \Delta\mathbf{L} \end{bmatrix} = 0$$
(3)

Equation (3) can be used to describe an arbitrary local motion at a given configuration of the manipulator [9]. When first order terms of Eq. (3) are sufficient to describe the motion, the manipulator is in a regular configuration and the following equation can be used instead of Eq. (3)

$$\frac{\partial \Gamma(\mathbf{L},\boldsymbol{\theta})}{\partial \boldsymbol{\theta}} \Delta\boldsymbol{\theta} + \frac{\partial \Gamma(\mathbf{L},\boldsymbol{\theta})}{\partial \mathbf{L}} \Delta\mathbf{L} = 0 \qquad (4)$$

Otherwise the configuration $(\mathbf{L}, \boldsymbol{\theta})$ is special and the manipulator meets a singularity. This happens when the constraint Jacobian $\partial \Gamma / \partial \boldsymbol{\theta}$ drops rank so that the second order terms of the equation (3) are needed to describe the constraints. The three vertices of the moving platform have the following coordinates in the fixed reference frame

$$\mathbf{b}_1 = \begin{bmatrix} L_1 \cos(\theta_1) & L_1 \sin(\theta_1) \end{bmatrix}^T$$
$$\mathbf{b}_2 = \begin{bmatrix} A_{2x} + L_2 \cos(\theta_2) & L_2 \sin(\theta_2) \end{bmatrix}^T$$
$$\mathbf{b}_3 = \begin{bmatrix} A_{3x} + L_3 \cos(\theta_3) & A_{3y} + L_3 \sin(\theta_3) \end{bmatrix}^T.$$

Thus, the constraint Jacobian can be put in the following form

$$\frac{\partial \Gamma}{\partial \boldsymbol{\theta}} = 2 \begin{bmatrix} L_1(A_{2x}s_1 + L_2 s_{12}) & L_2(L_1 s_{21} - A_{2x}s_2) & 0 \\ 0 & -L_2((A_{2x}-A_{3x})s_2 & L_3((A_{2x}-A_{3x})s_3 \\ & -L_3 s_{23} + A_{3y} c_2) & -L_2 s_{23} + A_{3y} c_3) \\ L_1(A_{3x}s_1 - L_3 s_{31} & 0 & -L_3(A_{3x}s_3 - L_1 s_{31} \\ -A_{3y} c_1) & & -A_{3y} c_3) \end{bmatrix}$$
(5)

where $s_i = \sin(\theta_i)$, $c_i = \cos(\theta_i)$ and $s_{ij} = \sin(\theta_i - \theta_j)$.

### III. SINGULAR CURVES IN 2-DIMENSIONAL SLICES OF THE CONFIGURATION SPACE

The singularities were determined in [9] by looking for the configurations where $\det(\partial \Gamma / \partial \boldsymbol{\theta})$ but this equation was incorrectly displayed. To derive the right equation, we have used a geometric approach that is much more direct than the calculation of the determinant, which does not simplify easily. The manipulator is in a singular configuration whenever the axes of its three leg rods intersect (possibly at infinity). The derivation of this geometric condition is straightforward and yields the following equation

$$A_{2x}s_2 s_{31} + (A_{3x}s_3 - A_{3y}c_3)s_{12} = 0, \qquad (6)$$

which we have checked successfully with several manipulator geometries. In order to reduce the dimension of the problem, [9] shows that it is possible to consider two-dimensional slices of the configuration space by fixing one of the leg rod lengths. By doing so, only two parameters are needed to fully define a configuration. For a fixed value of $L_1$, a configuration may be fully defined by either $(\alpha, \theta_1)$ or $(L_2, L_3)$. Note that in the first case, the configuration is defined in the output space by the position and the orientation of the moving platform ($L_1$ and $\theta_1$ define the position of $B_1$ in the plane and $\alpha$ defines the orientation of the moving platform in the plane). In the second case, the configuration is defined in the joint space by the three leg rod lengths.

Now, for any fixed value of $L_1$, it is possible to plot the singular curves in either $(\alpha, \theta_1)$ or $(L_2, L_3)$. But we first need to rewrite Eq. (6) as function of $L_1$, $\alpha$ and $\theta_1$, which we do with the help of the following set of geometric equations

$$\begin{cases} A_{2x} + L_2 c_2 - L_1 c_1 - b_1 \cos(\alpha) = 0 \\ L_2 s_2 - L_1 s_1 - b_1 \sin(\alpha) = 0 \\ A_{3x} + L_3 c_3 - L_1 c_1 - b_3 \cos(\alpha + \beta) = 0 \\ A_{3y} + L_3 s_3 - L_1 s_1 - b_3 \sin(\alpha + \beta) = 0, \end{cases} \quad (7)$$

where $\beta$ is the (fixed) angle between $B_1B_2$ and $B_1B_3$ (Fig. 1). The first (resp. last) two equations make it possible to express $L_2$ (resp. $L_3$) as function of $L_1$, $\alpha$ and $\theta_1$. Then, $c_2$ and $s_2$ (resp. $c_3$ and $s_3$) are calculated as function of $L_1$, $\alpha$ and $\theta_1$ from the first (resp. last) two equations of (7) and their expressions are input in Eq. (6), which, now, depend only on $L_1$, $\alpha$ and $\theta_1$. The resulting equation makes it possible to plot the singular curves directly in $(\alpha, \theta_1)$. To get the curves in $(L_2, L_3)$, we use the expressions of $L_2$ and $L_3$ as function of $L_1$, $\alpha$ and $\theta_1$.

Figure 2 shows the singular curves obtained for the 3-*RPR* manipulator of [8, 9]. The geometric parameters of this manipulator are recalled below in an arbitrary length unit:

$A_1=(0, 0)$      $d_1=17.04$
$A_2=(15.91, 0)$      $d_2=16.54$
$A_3=(0, 10)$      $d_3=20.84$

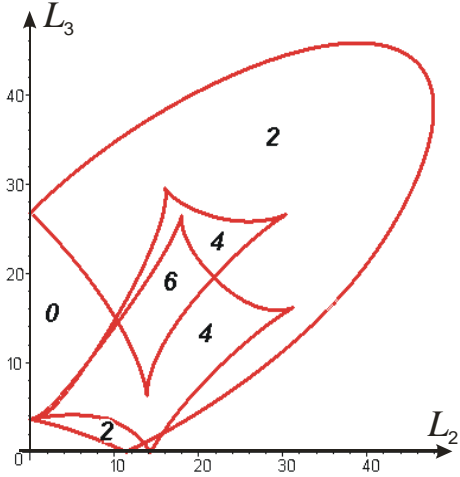

Figure 2. Singular curves in $(L_2, L_3)$ for $L_1=14.98$. Number of assembly modes is displayed in each region.

The singular curves were plotted in the slice of the joint space defined by $L_1=14.98$ as in [9]. These curves give rise to several regions with a constant number of assembly modes in each region. Figure 2 shows that several cusp points exist where three assembly modes coalesce. In [9] the number of cusp points was claimed to be five. It will be shown in next section that a sixth cusp point does exist.

## IV. DETERMINATION OF THE CUSP POINTS

For serial 3-DOF manipulators, the cusp points can be determined by deriving the condition under which the inverse kinematics polynomial admits three identical roots [11]. However this approach is intractable when applied to the direct kinematics polynomial of 3-*RPR* manipulators because the algebra involved is too complicated.

An interesting alternative approach was proposed in [9] by writing the condition under which the manipulator loses first and second order constraints. The resulting condition for triple coalescence of assembly modes was shown to take the following form

$$\mathbf{v}^T \left[ u_1 \frac{\partial^2 \Gamma_1}{\partial \boldsymbol{\theta}^2} + u_2 \frac{\partial^2 \Gamma_2}{\partial \boldsymbol{\theta}^2} + u_3 \frac{\partial^2 \Gamma_3}{\partial \boldsymbol{\theta}^2} \right] \mathbf{v} = 0, \quad (8)$$

where $\mathbf{v}$ is a unit vector in the right kernel of matrix $\partial \boldsymbol{\Gamma}/\partial \boldsymbol{\theta}$, and $u_1, u_2, u_3$ are the three components of the unit vector $\mathbf{u}$ that spans the left kernel. Vectors $\mathbf{u}$ and $\mathbf{v}$ can be chosen in the set of nonzero rows and columns of the adjoint of matrix $\partial \boldsymbol{\Gamma}/\partial \boldsymbol{\theta}$ (i.e. the matrix of cofactors of the transpose of $\partial \boldsymbol{\Gamma}/\partial \boldsymbol{\theta}$), respectively.

Calculating the adjoint of $\partial \boldsymbol{\Gamma}/\partial \boldsymbol{\theta}$ from Eq. (5) yields

$$\text{adj}\left(\frac{\partial \boldsymbol{\Gamma}}{\partial \boldsymbol{\theta}}\right) = \begin{pmatrix} k_1 k_2 & -k_2 k_5 & k_3 k_5 \\ k_3 k_4 & k_2 k_6 & k_3 k_6 \\ -k_1 k_4 & k_4 k_5 & k_1 k_6 \end{pmatrix} \quad (9)$$

Expressions for $k_1$ through $k_6$ were reported in [9] with several errors. The corrected expressions are given below:

$$\begin{aligned} k_1 &= 2L_2 \left( (A_{3x} - A_{2x}) s_2 + L_3 s_{23} - A_{3y} c_2 \right) \\ k_2 &= -2L_3 \left( L_1 s_{13} + A_{3x} s_3 - A_{3y} c_3 \right) \\ k_3 &= -2L_3 \left( (A_{3x} - A_{2x}) s_3 + L_2 s_{23} - A_{3y} c_3 \right) \\ k_4 &= 2L_1 \left( L_3 s_{13} + A_{3x} s_1 - A_{3y} c_1 \right) \\ k_5 &= -2L_2 \left( L_1 s_{12} + A_{2x} s_2 \right) \\ k_6 &= 2L_1 \left( L_2 s_{12} + A_{2x} s_1 \right) \end{aligned} \quad (10)$$

Taking $\mathbf{u}$ (resp. $\mathbf{v}$) as the first row (resp. column) of (9), Eq. (8) can be written as

$$\begin{pmatrix} k_1 k_2 & k_3 k_4 & -k_1 k_4 \end{pmatrix} \left( k_1 k_2 \frac{\partial^2 \Gamma_1}{\partial \boldsymbol{\theta}^2} - k_2 k_5 \frac{\partial^2 \Gamma_2}{\partial \boldsymbol{\theta}^2} + k_3 k_5 \frac{\partial^2 \Gamma_3}{\partial \boldsymbol{\theta}^2} \right) \begin{pmatrix} k_1 k_2 \\ k_3 k_4 \\ -k_1 k_4 \end{pmatrix} = 0 \quad (11)$$

where

$$\frac{\partial^2 \Gamma_1}{\partial \boldsymbol{\theta}^2} = 2 \begin{bmatrix} L_1(A_{2x}c_1 + L_2 c_{21}) & -L_1 L_2 c_{21} & 0 \\ -L_1 L_2 c_{21} & -L_2(A_{2x}c_2 - L_1 c_{21}) & 0 \\ 0 & 0 & 0 \end{bmatrix}$$

$$\frac{\partial^2 \Gamma_2}{\partial \boldsymbol{\theta}^2} = 2 \begin{bmatrix} 0 & 0 & 0 \\ 0 & \begin{matrix}-L_2((A_{2x}-A_{3x})c_2 \\ -L_3 c_{23} - A_{3y}s_2)\end{matrix} & -L_2 L_3 c_{23} \\ 0 & -L_2 L_3 c_{23} & \begin{matrix}L_3((A_{2x}-A_{3x})c_3 \\ +L_2 c_{23} - A_{3y}s_3)\end{matrix} \end{bmatrix}$$

$$\frac{\partial^2 \Gamma_3}{\partial \boldsymbol{\theta}^2} = 2 \begin{bmatrix} L_1(A_{3x}c_1 + L_3 c_{31} + A_{3y}s_1) & 0 & -L_1 L_3 c_{31} \\ 0 & 0 & 0 \\ -L_1 L_3 c_{31} & 0 & L_3(L_1 c_{31} - A_{3x}c_3 - A_{3y}s_3) \end{bmatrix}$$
(12)

In [9], Eq. (11) was left as such and no information was provided on how to use it in a computer program. We have developed an algorithm to solve this equation for any 3-*RPR* manipulator and we have implemented it in Maple. The resulting Maple program is available on our webpage (because of the anonymous review process, the url cannot be displayed in this proposal). Equation (11) is first expressed as function of $L_1$, $\alpha$ and $\theta_1$, which we do as in section III using Eqs. (7). Then, $\sin(\theta_1)$, $\cos(\theta_1)$, $\sin(\alpha)$ and $\cos(\alpha)$ are replaced by their expression in $t_1 = \tan(\theta_1/2)$ and $t = \tan(\alpha/2)$, respectively, in order to have pure algebraic expressions for Eqs. (6) and (11). At this stage, Eq. (6) (resp. Eq. (11)) results in a 4th-order (resp. 12th-order) polynomial in $t$ and $t_1$. Then, $t$ is eliminated from these two polynomials by computing their resultant [12]. Because Eq. (11) is complicated, this task requires careful preliminary algebraic manipulation. The resulting equation has the following general form

$$P_1^{a_1} P_2^{a_2} P_3^{a_3} \dots P_{n-1}^{a_{n-1}} P_n^{a_n} Q = 0 \quad (13)$$

where $Q$ is a 24th-order univariate polynomial in $t_1$ and $P_1$, $P_2$,…,$P_n$ are quadratic and quartic polynomials in $t_1$. It is well known that elimination often generate spurious solutions [12]. This is the case in Eq. (13) where $P_1$, $P_2$,…,$P_n$ can be shown to define no cusp points. Thus, only the 24th-order univariate polynomial $Q$ is relevant. Once the real roots of $Q$ are calculated, $t$ can be found by back-substitution in Eq. (6) and solving for the resulting quartic equation. Finally, the cusps points are obtained using the expressions of $L_2$ and $L_3$ as function of $L_1$, $\alpha$ and $\theta_1$.

The algorithm was first run for the manipulator studied in the previous section and for $L_1$=14.98 (Fig. 2). Six cusp points were identified instead of five [9]. These points are pinpointed with circles in Fig. 3. The sixth point missed in [9] is circled with bold lines and in-boxed in a separate view. As reported in [9], the points appearing on $L_2$=0 and $L_3$=0 are not cusp points because the constraint equations are non-differentiable.

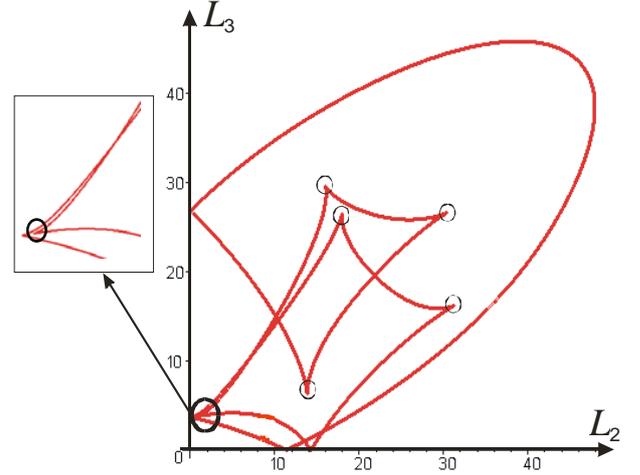

Figure 3. Singular curves in ($L_2$,$L_3$) for $L_1$=14.98. The six cusp points are circled, the one in-boxed in a separate view was missed in [9].

## V. DESCRIPTIVE ANALYSIS

In this section the singular curves are analysed in several slices of the joint space for the manipulator of [8, 9] whose geometric parameters were given in section III. Figure 4 depicts the singular curves for increasing values of $L_1$ and shows that the number of cusp points is not the same for all slices as we may have 0, 2, 4, 6 or 8 cusp points. In Fig. 4, regions with two assembly modes (resp. four, six) are filled in light grey (resp. in dark grey, in black). Zero-cusp slices are obtained for very small values of $L_1$ only ($L_1$=0.05 in Fig. 4), where the singular curves are made of two separate closed curves that define only one small region with two assembly modes and a large void (note, the two curves are so close that the region cannot be seen on Fig. 4). When $L_1$ is increased two cusp points appear, the void gets smaller and a four-solution region appears ($L_1$=2). Then two more cusp points appear, defining one more four-solution region ($L_1$=2.8). We have six cusp points and a small void at $L_1$=6; for $L_1$=8, 10, 12, 14, 16, 18, 20, 24 and 26, we have always six cusp points but the void is replaced with a six-solution region. Eight cusp points were found in a small vicinity of $L_1$=27. Then two cusp points and the six-solution region disappear ($L_1$=29). Finally, the number of cusp points stabilizes to four, defining one central four-solution region surrounded by a two-solution region ($L_1$=31, …). Interestingly, this last pattern is very similar to the one often observed in a cross-section of the workspace of 3-*R* serial manipulators [13]. However, serial manipulators feature the same pattern in all cross-sections (the sections which passes through the first revolute joint axis), and variation in the number of cusp points arises only from a modification of the manipulator geometry.

The above analysis shows that the joint space topology of 3-*RPR* manipulators is very complicated. Contrary to serial manipulators, the shape of the singular curves and the number of cusps points depend on which slice is chosen in the joint space. Thus, planning trajectories is not easy. However, we have noticed that the pattern stabilizes for sufficiently large values of $L_1$.

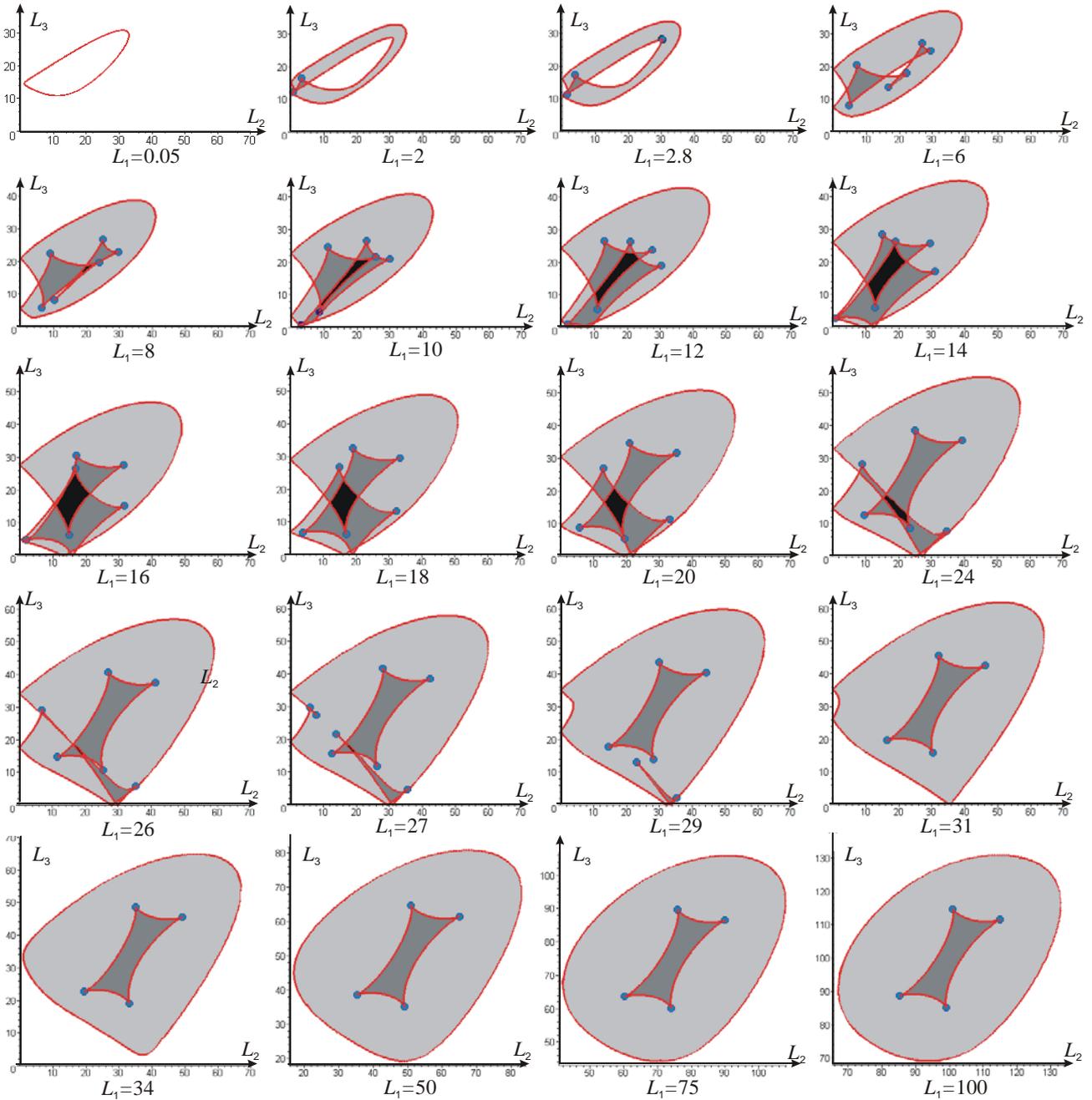

Figure 4. Singular curve patterns for increasing values of $L_1$.

This feature has been observed in many other manipulator geometries. For example, the 3-*RPR* manipulator defined by the following geometric parameters

$A_1=(0., 0.)$     $d_1=1.3$
$A_2=(3, 0.)$     $d_2=0.9$
$A_3=(1.1, 2.7)$     $d_3=0.4$

has a constant pattern as soon as $L_1>5$ (Fig. 5). Note that, in contrast with the stabilized pattern obtained for the preceding manipulator, this one features a large void.

Most research on parallel manipulators has been focused on the analysis and optimization of the workspace. If the workspace is useful for manipulator design, the analysis of singular curve patterns in the joint space can be used as a complementary tool to compare several manipulator geometries.

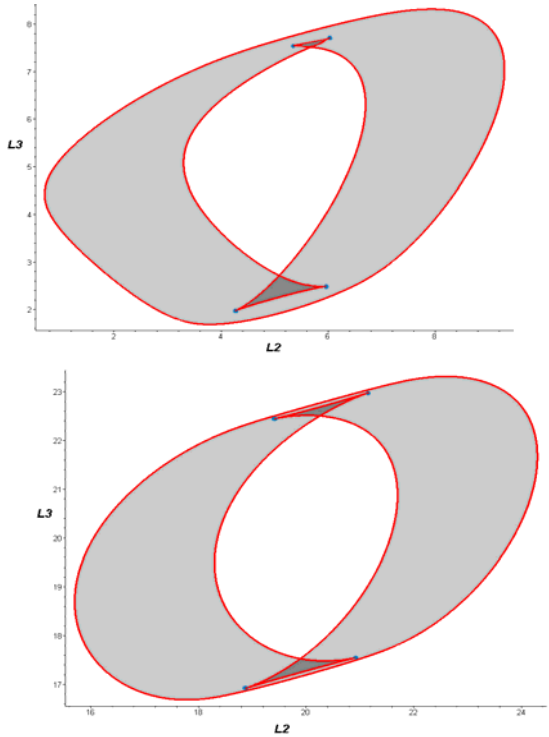

Figure 5.  Stabilized singular curve pattern for another manipulator geometry plotted for $L_1=5$ (up) and $L_1=20$ (down).

## VI. Conclusions

A descriptive analysis of the singular curves in slices of the joint space of 3-*RPR* parallel manipulators was conducted in this paper, with a special focus on the determination of cusp points on these singular curves. This work was based on a meticulous review of a previous theoretical work and the recalculation of some of its formulas that were not correctly displayed. The existence condition of triple assembly modes was exploited and an algorithm for the automatic detection of cusp points was developed and run for several manipulators with arbitrary geometries. It was shown that contrary to what arises in serial manipulators where any cross section of the workspace exhibits the same pattern of singular curves and cusp points, this pattern depends on the choice of the slice in the joint space for a given 3-*RPR* parallel manipulator. On the other hand, we have noticed that it is possible to have a constant pattern by adjusting the joint limits. It has been observed that the maximum number of cusps depends on the manipulator geometry but no manipulators were found with more than eight cusp points. Most research on parallel manipulators has been focused on the analysis and optimization of the workspace. This work is a complementary tool that helps better understand the topology of the joint space of parallel manipulators and finds applications in both design and trajectory planning.